\pdfoutput=1
\documentclass[11pt]{article}

\usepackage[]{ACL2023}

\usepackage{times}
\usepackage{latexsym}
\usepackage{graphicx}
\usepackage{enumitem}
\usepackage{pgfplots}
\pgfplotsset{compat=1.16}
\usepgfplotslibrary{statistics}
\usepackage[font=normalsize,labelfont={bf}]{caption}
\usepackage{algpseudocode}% http://ctan.org/pkg/algorithmicx
\usepackage[ruled,linesnumbered]{algorithm2e}
\usepackage{multirow}% http://ctan.org/pkg/multirow
\usepackage{booktabs}%
\usepackage[T1]{fontenc}
\usepackage[utf8]{inputenc}

\usepackage{microtype}

\usepackage{inconsolata}

\title{Replace and Report: NLP Assisted Radiology Report Generation}

\author{Kaveri Kale, Pushpak Bhattacharyya and Kshitij Jadhav \\
 Indian Institute of Technology Bombay \\
  \texttt{\{kaverikale,pb\}@cse.iitb.ac.in} and kshitij.jadhav@iitb.ac.in\\}

\begin{document}
\maketitle
\begin{abstract}
Clinical practice frequently uses medical imaging for diagnosis and treatment. A significant challenge for automatic radiology report generation is that the radiology reports are long narratives consisting of multiple sentences for both abnormal and normal findings. Therefore, applying conventional image captioning approaches to generate the whole report proves to be insufficient, as these are designed to briefly describe images with short sentences. We propose a template-based approach to generate radiology reports from radiographs. Our approach involves the following: i) using a multilabel image classifier, produce the tags for the input radiograph; ii) using a transformer-based model, generate pathological descriptions (a description of abnormal findings seen on radiographs) from the tags generated in step (i); iii) using a BERT-based multi-label text classifier, find the spans in the normal report template to replace with the generated pathological descriptions; and iv) using a rule-based system, replace the identified span with the generated pathological description. We performed experiments with the two most popular radiology report datasets, IU Chest X-ray and MIMIC-CXR and demonstrated that the BLEU-1, ROUGE-L, METEOR, and CIDEr scores are better than the State-of-the-Art models by 25\%, 36\%, 44\% and 48\% respectively, on the IU X-RAY dataset. To the best of our knowledge, this is the first attempt to generate chest X-ray radiology reports by first creating small sentences for abnormal findings and then replacing them in the normal report template. 
\end{abstract}

\section{Introduction}
Radiology report generation, which aims to automatically generate a free-text description of a clinical radiograph (like a chest X-ray), has become an important and interesting area of research in both clinical medicine and artificial intelligence. Natural Language Processing (NLP) can speed up report generation and improve healthcare quality and standardization. Thus, recently, several methods have been proposed in this area \citep{jing2017automatic, li2018hybrid, yuan2019automatic, chen2020generating, alfarghaly2021automated}. 
Radiology reports are lengthy narratives, which makes report generation difficult. Therefore, applying conventional image captioning approaches to generate the whole report \citep{vinyals2015show, anderson2018bottom} proves to be insufficient, as such approaches are designed to briefly describe images with short sentences. Further, even if benchmark datasets are balanced between normal and abnormal studies, multiple organs' findings are included in a single report, and if at least one organ is abnormal, the report is classified as abnormal, but it still contains more normal sentences than abnormal. As a result, existing text generation models may be overly focused on widely reported normal findings. Hence, we propose an approach to generating radiology reports by generating pathological descriptions, i.e., descriptions of abnormal findings, and then replacing corresponding normal descriptions from the normal report template with a generated pathological description.
Experimental results on the IU Chest X-ray (referred to as IU X-RAY) and the MIMIC-CXR benchmark datasets confirm the validity and effectiveness of our approach and demonstrate that our approach achieves better results than the State-of-the-Art methods. Two experts in the field were involved in this work. One is a radiologist with 30 years of experience, and the other is a doctor with an MBBS, MD, and Ph.D. and 2 years of experience in medicine.

Our contributions are:
\begin{enumerate}[noitemsep]
    \item A new approach to generating radiology reports: i) generating the tags for the input radiograph; ii) generating pathological descriptions from the tags generated in step (i); iii) identifying the spans in the normal report template to replace with the generated pathological descriptions; and iv) replacing the identified span with the generated pathological description—improves the quality and accuracy of the radiology reports. Compared to the previous State-of-the-Art models, the BLEU-1, ROUGE-L, METEOR, and CIDEr scores of the pathological descriptions generated by our approach are raised by 25\%, 36\%, 44\% and 48\% respectively, on the IU X-RAY dataset.
    \item A dataset of tags (tags are the disease keywords and radiological concepts associated with X-ray images) and their corresponding pathological descriptions. (Derived from IU X-RAY and MIMIC-CXR datasets containing 3827 and 44578 data points respectively.)
    \item A dataset of the pathological descriptions and the corresponding normal sentences from the normal report template to replace with pathological descriptions (6501 data points).
\end{enumerate}

\section{Related Work}
The topic of automatic report generation was researched by \citet{jing-etal-2018-automatic, zhang2017mdnet,yuan2019automatic}. Several attempts have been made in the medical field to create medical reports from the corresponding images. Most researchers use multilabel image captioning to produce X-ray reports, and they subsequently use those captions as textual features. The IU X-ray dataset was created by \citet{demner2016preparing} to generate radiology reports automatically. The IU X-RAY dataset's chest X-ray images were used to generate the first structured report using tags predicted by a CNN-RNN model \cite{shin2016learning}. A system for generating natural reports for the Chest-Xray14 dataset, employing private reports, was presented by \citet{wang2017chestx}. This framework used a non-hierarchical CNN-LSTM architecture and focused on semantic and visual aspects. Visual attention given to recurrent decoders and convolution-recurrent architectures (CNN-RNN) was first introduced by \citet{vinyals2015show} on image captioning. 

Radiology report generation has recently shifted to transformer-based models \cite{vaswani2017attention, devlin2018bert}. Knowledge-driven Encode, Retrieve, and Paraphrase (KERP) \cite{li2019knowledge} is used for accurate and reliable medical report generation. \citet{yuan2019automatic} suggests pretraining the encoder with several chest X-ray images to properly recognise 14 typical radiographic observations. According to \citet{chen2020generating} proposals, radiology reports can be generated using a memory-driven transformer, while \citet{pino2021clinically} suggests a template-based X-ray report generation approach. \citet{pino2021clinically} clinically defines the templates for each abnormality to indicate its presence or absence. If the generated tags indicate any disease, then the system will choose the corresponding abnormal sentence template. This method cannot generate patient-specific data like anatomical location, size, \textit{etc}. \citet{wang2021tmrgm} proposed a template-based multi-attention report generation model (TMRGM) for normal and abnormal studies. This work uses template-based techniques exclusively for normal reports. 

The differences between our work and previous work are as follows: i) Instead of generating the whole report at once, we generate smaller sentences for only abnormal findings and then replace them in the normal report template. ii) Unlike other state-of-the-art models, our methodology does not put excessive emphasis on normal sentences. iii) If the report's findings are normal, we use a standard template for normal reports.

\section{Methodology}
The approach that we follow to generate the radiology reports from the radiographs is as follows:

\begin{itemize}[noitemsep]
    \item Generate the tags for the input chest radiograph.
    \item Generate the pathological description from generated tags.
    \item Replace appropriate normal sentences (referred to as \textbf{normal description}) from the normal report template with the generated pathological descriptions.
    \begin{itemize}
        \item Identify the span in the normal report template to replace.
        \item Replace the identified span with the generated pathological description.
    \end{itemize}
\end{itemize}

\subsection{Model Architecture}
\begin{figure*}[htb!]
\centering
\includegraphics[width=\textwidth]{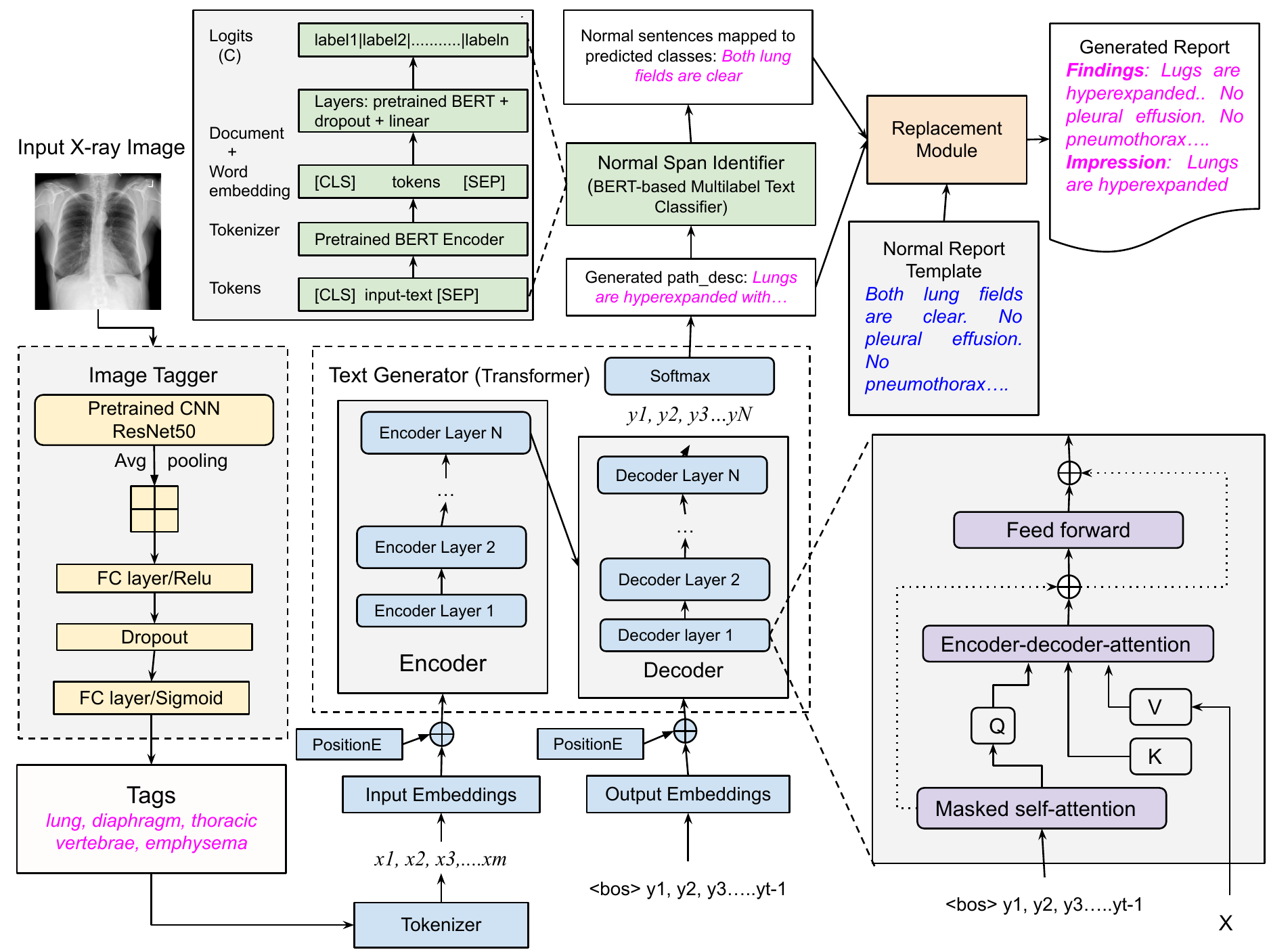}
\caption{The architecture of our proposed model: Our model has four important components: an image tagger, a text generator, a span identifier, and a replacement module. The image tagger module produces tags for the input X-ray image. The text-generator module generates the pathological description for input tags. The span identifier module identifies the normal sentences that need to be replaced by generated pathological descriptions in the normal report template. The replacement module replaces identified normal spans with generated pathological descriptions.}
\label{fig:model_arch}
\end{figure*}

Figure \ref{fig:model_arch} illustrates the four main components of our model, which are the image tagger, text-generator (i.e., transformer), span identifier, and replacement module. The overall description of the four parts is provided below.

\subsubsection{Image Tagger}
Tagging X-ray images with multiple tags is a multi-label image classification problem. We build our multi-label classifier on top of convolutional neural network (CNN) architectures such as ResNet-50. Our model takes an X-ray image as input, calculates a score for each of the \(L\) labels, and then uses a cutoff threshold to decide which labels to keep.
Target in each sample presented a binary vector \([{y_1}^j, {y_2}^j,..., {y_L}^j], j = 1...n, y \in Y\) and indicates whether a label is present (1) or absent (0). We use the sigmoid activation function in the output layer, and the binary cross-entropy loss function is used to fit the model. The loss is calculated as: 
\begin{equation}
{Loss = \frac{1}{L}\sum_{i=1}^{n}[y_i log(\hat{y_i}) + (1 - y_i) log(1 - \hat{y_i})]}
\end{equation}
where \(\hat{y_i}\) is the \(i\)th predicted value by model, \(y_i\) is the corresponding target value, and \(L\) is the number of scalar values in the model output.
 
\subsubsection{Tags Embedding}
Input tag set \(X\) is tokenized as \(\{x_1, x_2, . . ., x_{|s|} \}\). Input token embeddings is given by, \(e_x = \{e_{x_1}, e_{x_2}, . . ., e_{x_{|n|}} \}\). For a token \(x_i\), its embedding is \(e_{x_i} \in R^d\), where \(d\) is the dimension of the token embeddings. The positional encoding is added to the input embeddings before passing it to the transformer layer. The text embeddings are the sum of the token embeddings and the positional embeddings, i.e., \(e_{xp} = e_x + e_p \), where \(e_p\) is the positional embeddings.

\subsubsection{Encoder-decoder Architecture}
Generating pathological descriptions from tags can be looked at as a text generation problem. Given the input tag sequence \(X = (x_1, x_2,...,x_s)\), the goal of general text generation model is to generate the pathological description sequence \(Y = (y_1, y_2,...,y_T)\), where \(S\) and \(T\) are the length of the input and output sequence respectively. The text generation model can be defined as follows: 
\begin{equation}
    p(Y|X) = \prod\limits_{t=1}^Tp(y_t|X,Y_{<t})
\end{equation}

Finally, it generates a pathological description with the highest probability. If the image is labeled as \textit{normal} then the text generation module generates the sentence \textit{ No acute abnormality found.} Table \ref{fig:data_sample} shows the samples from the dataset that include tags and their corresponding pathological descriptions.

\subsubsection{Span Identification and Replacement Module}
\begin{figure}[htb!]
\centering
\includegraphics[width=\columnwidth]{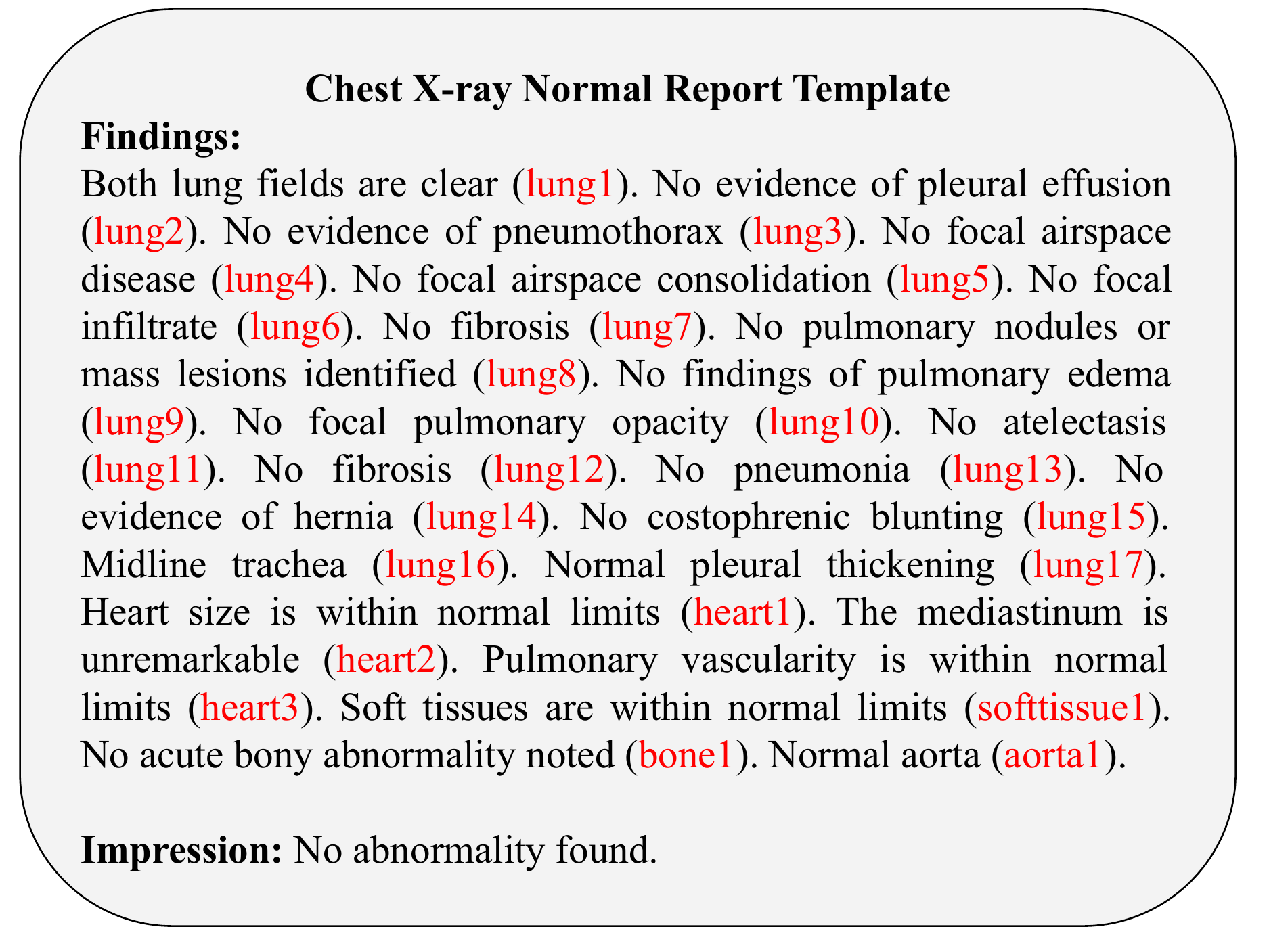}
\caption{Chest X-ray normal report template. Each sentence is mapped with a unique label. Labels are highlighted in red.}
\label{fig:normalReport}
\end{figure}

\paragraph{Span Identification:} The span identification module identifies the span from the normal report template that would be replaced with a generated pathological description. To create a standard normal report template, we manually curated a set of sentences per abnormality indicating absence of abnormality (that means normal findings), totaling 23 sentences. With the expert's opinion, we built the template by examining the reports and picking existing sentences or creating new ones. Due to the fixed nature of sentences, we treat each as a separate label. Figure \ref{fig:normalReport} shows an example of the normal report template with a label for each sentence mentioned in brackets. We create a dataset of pathological description sentences, which we extract from the findings and impressions section of the original dataset, and annotate it with corresponding normal sentence labels (i.e., \textit{lung1}, \textit{lung2}, \textit{etc.}). If a pathological description cannot be replaced with any of the normal sentences, then we annotate that sentence with the \textit{extra} label.  

This normal sentence identification problem can be formulated as a multilabel text classification problem. Consider input sentence as series of words: \(X = \{w_1, w_2,..., w_n\}\) and for span identification, we need to predict the sentence category \(Y = \{lung1, lung2,....,extra\}\).
We use a BERT-based multilabel text classifier to identify the normal sentences. The last layer uses a sigmoid activation function to generate the probability of a sample belonging to the corresponding classes. The loss function is the same as equation 1.

\paragraph{Replacement:} Once we obtain the list of normal sentences to replace with the generated pathological description, we submit the normal sentences, the normal report template, and the generated pathological description to the replacement module. If a sentence is labeled as \textit{extra} then we do not replace any normal sentences, but instead add pathological description sentences as an extra part in the report. For example, if the generated pathological description sentence is, \textit{Multiple surgical clips are noted.}, then we add this as an extra part in the report without removing any normal sentences. Replacement module finds the exact match of the identified normal span sentences with normal report sentences and replaces it with generated pathological description. If the generated description is \textit{No acute abnormality found.} then our replacement algorithm returns the standard normal report template as the generated report. If the span identifier gives multiple sentences to replace by a single sentence of pathological description, then the replacement module replaces the first sentence from the span and removes the remaining span sentences from the template. For example, if generated pathological description is \textit{Stable cardiomegaly with large hiatal hernia.}, then we have to replace \textit{No evidence of hernia} (\textit{lung14}) and \textit{Heart size is within normal limits.} (\textit{heart1}). Here, our replacement algorithm replaces \textit{lung14} sentence with a generated pathological description and removes the \textit{heart1} from the template. 

\section{Experiments}

In this section we cover the datasets, evaluation metrics and baselines used for the training and evaluation of our model in detail.
\subsection{Datasets}
We conduct our experiments on two datasets, which
are described as follows:
\begin{itemize}[noitemsep]
    \item \textbf{IU X-RAY} \cite{demner2016preparing} : a public radiography dataset compiled by Indiana University with 7,470 chest X-ray images and 3,955 reports. Each report has three parts: an impression, which is a title or summary of the report; findings, which contain the report in detail; and manual tags.
    \item \textbf{MIMIC-CXR} \cite{johnson2019mimic} : the largest publicly available radiology dataset that consists of
473,057 chest X-ray images and 227,943 reports. For the purpose of our experiments we utilized 44578 reports.
\end{itemize}

\begin{figure}[htb!]
\centering
\includegraphics[width=\columnwidth]{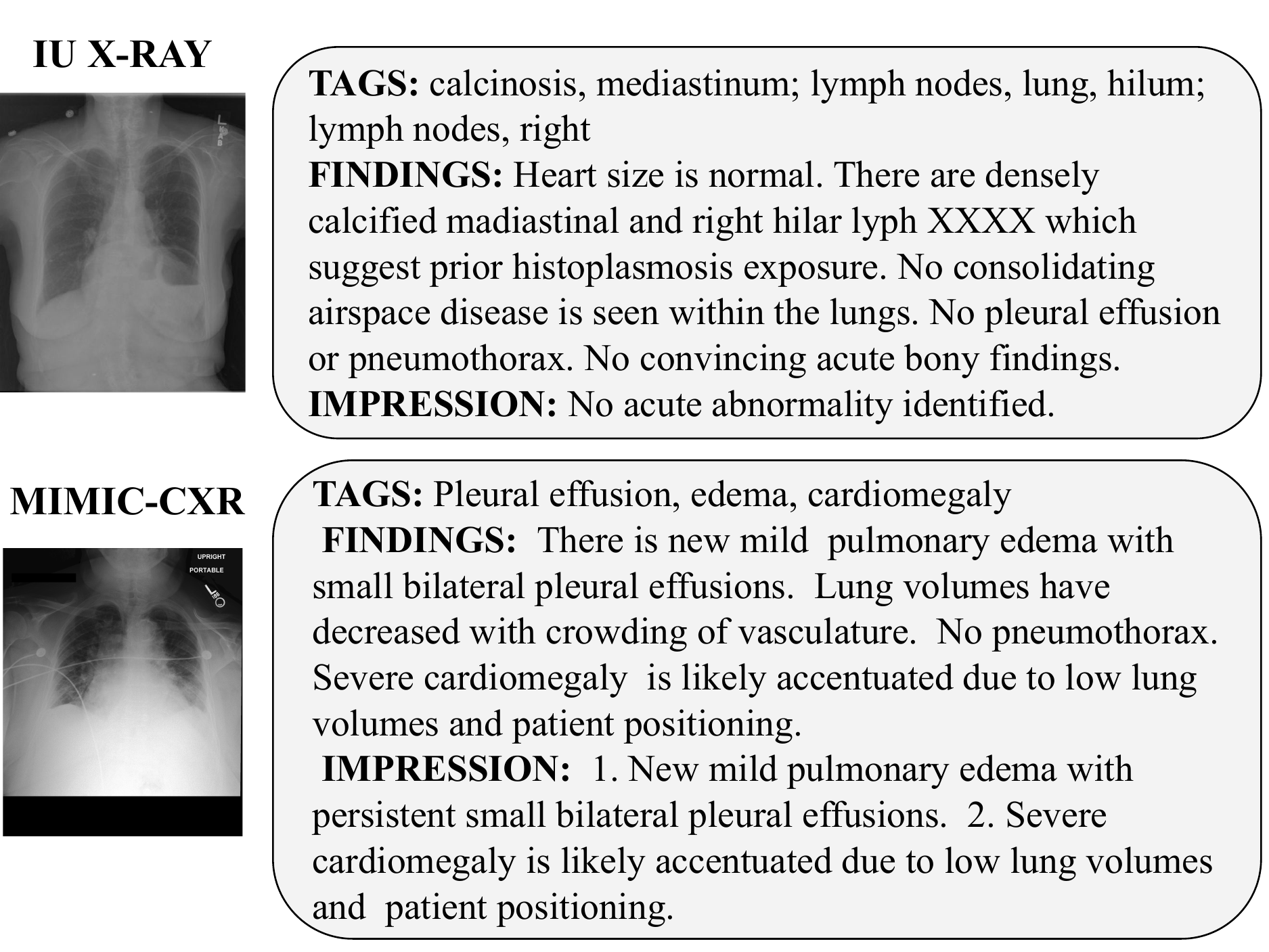}
\caption{The data samples are from the IU X-ray and MIMIC-CXR datasets.}
\label{fig:data_sample}
\end{figure}

Figure \ref{fig:data_sample} shows the data samples from the IU-Chest X-ray dataset and the MIMIC-CXR dataset.

We use distinct fields from the original datasets to train different modules. In this section, we give details about experimental settings for each module separately.
\subsubsection{Image Tagger Dataset} 
\begin{figure}[htb!]
\centering
\includegraphics[width=\columnwidth]{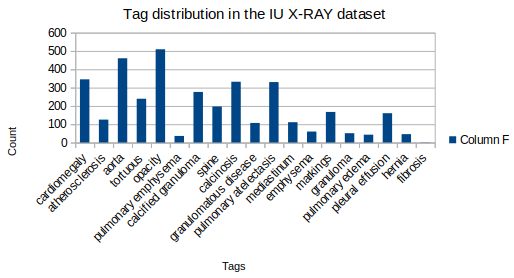}
\caption{IU X-RAY dataset tag distribution. There are a total of 189 unique tags but here we represent only 19 important tags.}
\label{fig:iu_class_dist}
\end{figure}

\begin{figure}[htb!]
\centering
\includegraphics[width=\columnwidth]{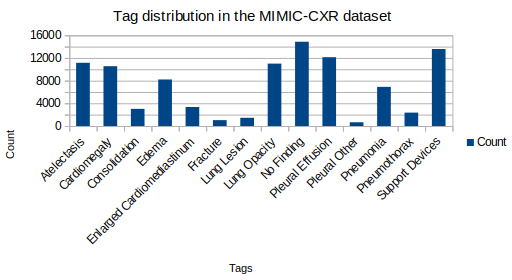}
\caption{MIMIC-CXR dataset tag distribution with all 14 tags.}
\label{fig:class_dist}
\end{figure}

We consider frontal chest-radiographs as input and target as tags for the IU X-RAY dataset and CheXpert labels for MIMIC-CXR dataset. We build a model using Convolutional Neural Networks (CNN) to analyze each image and classify it with one or more of the 189 labels for IU X-Ray dataset and 14 labels for MIMIC-CXR dataset. Figure \ref{fig:iu_class_dist} and figure \ref{fig:class_dist} shows the distribution of important classes in IU X-RAY dataset and MIMIC-CXR datasets, respectively.
For the IU X-RAY dataset out of 3827 samples train, validation and test split is 3000, 327 and 500 respectively. For the MIMIC-CXR dataset out of 44578 samples train, validation and test split is 40000, 2000 and 2578 respectively.

\subsubsection{Pathological Description Generator Dataset} 
\begin{table*}[hbt!]
\centering
\resizebox{\textwidth}{!}{%
\begin{tabular}{p{0.2\textwidth} p{0.5\textwidth} p{0.3\textwidth}} 
\hline
 \textbf{\small{Tags}} & \textbf{\small{Findings from the Original Dataset}} & \textbf{\small{Extracted Pathological Descriptions}} \\
\hline
\small{calcinosis, abdomen, left, severe} & \small{The heart size and cardiomediastinal silhouette are stable and within normal limits. Pulmonary vasculature appears normal. There is no focal air space consolidation. No pleural effusion or pneumothorax. Extensive left upper quadrant splenic calcification may reflect old granulomatous disease} & \small{Extensive left upper quadrant splenic calcification may reflect old granulomatous disease} \\
\hline
\small{nodule, lung, base, calcinosis, lung, hilum, lymph nodes, right, granuloma, right} & \small{There is a 1 cm nodule within one of the lung bases, seen only on the lateral view. There is a calcified right hilar lymph node and right granuloma. Heart size is normal. No pneumothorax.} & \small{There is a 1 cm nodule within one of the lung bases, seen only on the lateral view. There is a calcified right hilar lymph node and right granuloma.} \\
\hline
 \small{opacity, lung, apex, right, focal, opacity, lung, base, left, mild, spine, degenerative, cicatrix, lung, base, left, mild, pulmonary atelectasis, base, left, mild} & \small{The heart, pulmonary XXXX and mediastinum are within normal limits. There is no pleural effusion or pneumothorax. There is no focal air space opacity to suggest a pneumonia. There is a 1 cm focal opacity in the right lung apex incompletely evaluated by this exam. There is minimal left basilar XXXX opacity compatible with scarring or atelectasis. There are degenerative changes of the spine.} & \small{There is a 1 cm focal opacity in the right lung apex incompletely evaluated by this exam. There is minimal left basilar XXXX opacity compatible with scarring or atelectasis. There are degenerative changes of the spine.}  \\

\hline
\end{tabular}
}
\caption{The samples are from the IU X-ray dataset, including the findings and the pathological descriptions extracted from it.}
\label{table:notes}
\end{table*}
To train a model we derive two separate datasets from the original IU X-RAY and MIMIC-CXR datasets.
Two tasks are involved in creating this dataset:
\begin{itemize}[noitemsep]
    \item We constructed one more dataset that includes unique sentences from the IU X-RAY dataset, and we annotated them as \textit{normal} or \textit{abnormal}. The binary classifier dataset contains 5000 samples, each labeled as \textit{normal} or \textit{abnormal} by domain experts. We then train a BERT-based binary classifier to classify each sentence as \textit{normal} or \textit{abnormal}.
    \item We use the trained model to classify each sentence from findings and impressions into \textit{normal} and \textit{abnormal} classes. We remove normal sentences for each report and consider only abnormal sentences as our pathological description.
\end{itemize}
 Table \ref{table:notes} shows the examples of findings from the IU X-RAY dataset and extracted pathological descriptions from it.
The train, validation, and test split is the same as the image tagger dataset.

\subsubsection{Normal Span Identifier Dataset}

\begin{table}[hbt!]
\centering
\resizebox{\columnwidth}{!}{%
\begin{tabular}{ll} 
\hline
\textbf{Concept} & \textbf{Annotation labels} \\
\hline
cardiomegaly & {heart1} \\ 
{heart size} & {heart1} \\ 
{hilar} & {heart2} \\ 
{sternotomy} & {bone1} \\ 
{kyphosis} & {bone1} \\ 
{scoliosis} & {bone1} \\ 
{pleural fuild} & {lung2} \\ 
{atelectasis} & {lung1, lung11} \\ 
{consolidation} & {lung1, lung5, lung13} \\ 
{fibrosis} & {lung1, lung7} \\ 
{penumonia} & {lung1, lung13} \\ 
{costophrenic} & \multirow{2}{*}{lung2, lung15} \\ 
{blunting} & \\
{bronchial} & {lung1} \\ 
{granuloma} & {lung1 } \\ 
{COPD} & {lung1, lung4} \\ 
{interstitial} \\{marking} & {lung1} \\ 
\hline
\end{tabular}
\begin{tabular}{ll} 
\hline
\textbf{Concept} & \textbf{Annotation labels} \\
\hline

{airspace disease} & {lung1, lung4} \\ 
{infiltrate} & {lung1, lung6} \\ 
{nodule} & {lung1, lung8} \\ 
{pulmonary edema} & {lung1, lung9} \\ 
{clavicle}  & {bone1} \\
{shoulder}  & {bone1} \\
{humerous}  & {bone1} \\
{sternotomy} & {bone1} \\ 
{spine} & {bone1} \\ 
{bronchial cuffing} & {lung1, lung9} \\ 
{bronchovascular} & \multirow{2}{*}{lung1, lung11} \\ 
{crowding} & \\ 
{degenerative} & \multirow{2}{*}{bone1} \\ 
{changes} &  \\ 
{CABG} & {heart1} \\ 
{scarring} & {lung1, lung6, lung7} \\ 
{interstitial} & \multirow{2}{*}{lung1, lung6, lung7} \\
{prominence} & \\
\hline
\end{tabular}
}
\caption{Any mention of a concept given in column one in a pathological sentence should be labeled with the labels given in column two.}
\label{tbl:rules}
\end{table}

We construct a dataset to identify the sentences to replace from the normal report template with the generated pathological description. The constructed dataset contains sentences from the findings and impressions of the IU X-RAY dataset and their corresponding list of normal sentences to replace. 
Annotation guidelines are provided by the domain expert. Table \ref{tbl:rules} shows the guidelines provided by experts. Using those guidelines, we annotate the pathological description sentences. Data annotated by us is verified by a domain expert and corrected if necessary. Table \ref{table:multilabel_data} shows the samples from the span identification dataset. Out of total 6500 samples, the train, validation, and test splits are 5000, 500, and 1000, respectively. Table \ref{table:multilabel_data} shows the samples from the multilabel text classification dataset that we have constructed.

\begin{table}[hbt!]
\centering
\resizebox{\columnwidth}{!}{%
\begin{tabular}{p{0.75\columnwidth} p{0.25\columnwidth}} 
\hline
 \textbf{\small{Pathological Description}} & \textbf{\small{Labels}} \\
\hline
\small{The thoracic aorta is tortuous and calcified.} & \small{aorta1} \\
\hline
\small{XXXX right pleural opacity along the lower chest wall compatible with thickening and/or some loculated effusion, accompanied with some adjacent atelectasis / airspace disease within the right lung base.} & \small{lung2, lung4, lung10, lung11, lung17} \\
\hline
\small{Stable cardiomegaly with large hiatal hernia.}	& \small{lung14, heart1} \\
\hline
\small{Left greater than right basilar opacity, probable atelectasis and/or scarring.} & \small{lung10, lung11} \\
\hline
\end{tabular}
}
\caption{The samples are from the span identifier dataset. It includes the pathological descriptions and labels of the corresponding normal sentences to replace.}
\label{table:multilabel_data}
\end{table} 

\begin{table*}[hbt!]
\centering
\resizebox{\textwidth}{!}{%
\begin{tabular}{l|l|lllllll|lll}\hline
  \multirow{2}{*}{Dataset} & \multirow{2}{*}{Method} &\multicolumn{7}{c}{NLG Metrics} & \multicolumn{3}{c}{CE Metrics} \\ \cline{3-9} \cline{10-12}
  & &  \textbf{Bleu-1} & \textbf{Bleu-2} & \textbf{Bleu-3} & \textbf{Bleu-4}  & \textbf{Rouge-L} & \textbf{Meteor} & \textbf{CIDEr} & P & R & F\\
  \midrule
  \multirow{9}{*}{IU X-RAY} & CDGPT2 \citep{alfarghaly2021automated} & 0.387 & 0.245 & 0.166 & 0.111 & 0.289 & 0.164 & - & 0.0 & 0.0 & 0.0\\
    & Visual Transformer \citep{chen2020generating} & {0.470} & 0.304 & 0.219 & 0.165 & 0.371 & 0.187 & - & - & - & -\\ 
    & MvH+AttL+MC \citep{yuan2019automatic} & \underline{0.529} & \underline{0.372} & \underline{0.315} & \underline{0.255} & 0.453 & 0.343 & - & - & -\\
    & SentSAT+KG \citep{zhang2020radiology} & 0.441 & 0.291 & 0.203 & 0.147 & 0.367 & - & 0.304 & 0.483 & 0.490 & 0.478\\ 
    & CoAttn \citep{jing-etal-2018-automatic} & 0.517 & 0.386 & 0.306 & 0.247 & 0.447 & 0.217 & 0.327 & 0.491 & 0.503 & 0.491\\
     \cline{2-12}
    & CNN-TRG \citep{pino2021clinically} & 0.273 & - & - & - & 0.352 & - & 0.249 & \underline{0.529} & \underline{0.534} & \underline{0.540}\\   
    & TMRGM \citep{wang2021tmrgm} & 0.419 &	0.281 &	0.201 &	0.145 & 0.280 & 0.183 & 0.359\\
    \cline{2-12}
    & Ours (pathological description) & 0.402 & {0.322} & {0.285} & {0.180} & \underline{0.567} & \underline{0.455} & \underline{0.473} & \textbf{0.892} & \textbf{0.890} & \textbf{0.889} \\
    & Ours (full report) & \textbf{0.775} & \textbf{0.699} & \textbf{0.658} & \textbf{0.627} & \textbf{0.817} & \textbf{0.782} & \textbf{0.835} & 0.533 & 0.874 & 0.648
\\
  \midrule
   \multirow{5}{*}{MIMIC-CXR} & Visual Transformer \citep{chen2020generating} & 0.353 & 0.218 & 0.145 & 0.103 & 0.277 & 0.142 & - & 0.333 & 0.273 & 0.276\\   
    
    & SentSAT+KG \citep{zhang2020radiology} & \underline{0.441} & \underline{0.291} & \underline{0.203} & 0.147 & \underline{0.367} & - & 0.304 & - & - & -\\ 
    \cline{2-12} 
    & CNN-TRG \citep{pino2021clinically} & 0.080 & - & - & - & 0.151 & - & 0.026 & \underline{0.668} & \underline{0.749} & \underline{0.640}\\
    \cline{2-12}
    & Ours (pathological description) & 0.253 & {0.188} & {0.169} & \underline{0.163} & {0.348} & \underline{0.268} & \underline{0.331} & \textbf{0.769} & 0.771 & \textbf{0.765}\\
    & Ours (full report) & \textbf{0.833} & \textbf{0.807} & \textbf{0.794} & \textbf{0.785} & \textbf{0.833} & \textbf{0.861} & \textbf{0.861} & 0.488 & \textbf{0.863} & 0.606\\

    \hline
\end{tabular}
}
 \caption{The NLG metrics and CE metrics score of generated X-ray reports by previous methods and our approach vs. gold standard X-ray reports. The best results are in bold font, and the second best is underlined. }
  \label{tab:bleu}
\end{table*}

\subsection{Training Details}

We adopted the ResNet50 pretrained on Imagenet as the image tagger to produce tags for radiographs. We used frontal radiographs for both the IU X-RAY and the MIMIC-dataset. The IU X-RAY dataset contains 189 labels, and the MIMIC-CXR dataset contains 14 labels. According to the number of labels, the last layer of the image tagger module is different for both modules. We resize the images from both datasets to 224 x 224. We trained our model up to 20 epochs for both datasets. DGX A100-SXM-80GB GPU server takes 100 minutes for a single epoch for the MIMIC-CXR dataset and approximately 3 minutes for the IU X-RAY dataset.
The transformer in our proposed model was initialized with pretrained T5-large model weights. The model was trained under cross entropy loss with the ADAM optimizer. We set the learning rate to 1e-4. We decayed such a rate by a factor of 0.8 per epoch for each dataset and set the beam size to 5 to balance the generation's effectiveness and efficiency. Maximum input and output sequence lengths were set to 100. We train the transformer model up to 20 epochs for IU X-RAY dataset and up to 15 epochs for the MIMIC-CXR dataset. For one epoch it takes approximately 10 minutes and 60 minutes on a single DGX A100-SXM-80GB GPU for the IU X-RAY and MIMIC-CXR datasets, respectively. 
The last module was a BERT-based multilabel classifier, which identifies the normal span to replace. We used pretrained BERT weights to initialize our model. There are a total of 24 labels, according to the number of nodes in the last layer change. We train all the models on a DGX A100-SXM-80GB GPU server. For all transformer based models we use hugging face transformer libraries.\footnote{\url{https://huggingface.co/docs/transformers/index}}

\subsection{Evaluation}

We evaluate the results of different modules separately. For the image tagger and span identifier modules, we compare the accuracy, AUC score, precision, recall, and F1 scores. The first slot in the table \ref{tab:auc_result} shows the results for the image tagger module, and the second slot shows the results for the span identifier module on the IU X-RAY and MIMIC-CXR datasets.

\begin{table}[hbt!]
\centering
\resizebox{\columnwidth}{!}{%
\begin{tabular}{lllllll} 
\hline
\textbf{\small{Module}} & \textbf{\small{Dataset}} & \textbf{\small{Acc}}  & \textbf{\small{auROC}}  & \textbf{\small{F1 score}}  & \textbf{\small{Prec}}  & \textbf{\small{Recall}} \\
\hline
 \multirow{2}{*}{Image Tagger} & {MIMIC-CXR} & 0.71 & 0.82 & {0.68} & {0.80} & 0.64 \\
  & {IU X-RAY} & {0.75}  & {0.79} & {0.61}  & {0.71}  & {0.58} \\
\hline
  \multirow{2}{*}{Span Identifier} & {MIMIC-CXR} & 0.94 & 0.95 & 0.96 & 0.95 & 0.96 \\
   & {IU X-RAY} & 0.96  & 0.95 & 0.96  & 0.96  & 0.97 \\
\hline
\end{tabular}
}
\caption{Results of the image tagger module and span identification module on the IU X-RAY and MIMIC-CXR datasets given by our model. First slot shows the results for image tagger and second slot shows the results for span identifier.}
\label{tab:auc_result}
\end{table}

We compare the performance of our model with previous State-of-the-Art image captioning based methods like CNN-RNN \cite{vinyals2015show}, CDGPT2 \cite{alfarghaly2021automated} and Visual Transformer \cite{chen2020generating} and template based methods such as CNN-TRG \citep{pino2021clinically} and TMRGM \citep{wang2021tmrgm}.
To evaluate the generated pathological descriptions, we consider the pathological descriptions that we extract from original reports as ground truth. To evaluate the generated full reports, we generate templated reports by replacing ground truth pathological description in normal report template and consider it as ground truth.
The performance of the aforementioned models is evaluated by conventional natural language generation (NLG) metrics and clinical efficacy (CE) metrics. Clinical Efficacy (CE) metrics provides a quantitative assessment of the quality of generated radiology reports. Clinical efficacy (CE) metrics are calculated by comparing the critical radiology terminology extracted from the generated and reference reports. We use MIRQI\footnote{\url{https://github.com/xiaosongwang/MIRQI}}  tool to calculate the precision, recall, and F1 scores to evaluate the model performance for these metrics. NLG metrics such as BLEU \cite{papineni2002bleu}, ROUGE \cite{lin2004rouge}, CIDEr and METEOR \cite{banarjee2005}, which primarily focus on measuring n-gram similarities. For CE metrics, precision, recall, and F1-score are used to evaluate model performance. We use the CheXpert \cite{irvin2019chexpert} to label the generated reports and compare the results with ground truths in 14 different categories of thoracic diseases and support devices. Table \ref{tab:bleu} shows NLG and CE metrics of the generated pathological description as well as full reports by our model and baseline models compared to gold standard reports.

\subsection{Qualitative Evaluation and Error Analysis}
This section provides a qualitative analysis performed by a domain expert. The domain expert classified the generated reports into three categories: \textbf{accurate} (reports with most of the vital information), \textbf{missing details} (reports with no false information but missing some vital details), and \textbf{misleading} (reports with false information and an overall incorrect diagnosis). 
\begin{table}[hbt!]
\centering
\resizebox{\columnwidth}{!}{%
\begin{tabular}{ll@{\qquad}cccc}
  \toprule

  \textbf{Method} & \textbf{Samples} & \textbf{Accurate} & \textbf{Missing Details} & \textbf{Misleading}\\
  \midrule
  & $All (500)$ & \textbf{78.00}\% & \textbf{12.00}\% & \textbf{10.00}\%  \\
  Ours  & $Normal (183) $ & \textbf{99.95}\% & 00.00\% & \textbf{00.05}\% \\
    & $Abnormal (317)$ & \textbf{63.13}\% & \textbf{22.00}\% & \textbf{14.05}\% \\
    \midrule
    & $All (500)$	& 61.60\% & 28.20\% & 10.20\% \\
    CDGPT2  & $Normal (201)$	& 99.00\% & 00.00\% & 01.00\% \\
     & $Abnormal (299)$ & 36.50\% & 47.10\%	& 16.40\% \\
  \bottomrule
\end{tabular}
}
 \caption{Results of generated reports, manually evaluated by radiologist. Manual evaluation is done on the IU X-RAY dataset. Best results are shown in a bold face.}
  \label{tab:human_eval}
\end{table}
\begin{figure*}[htb!]
\centering
\includegraphics[width=\textwidth]{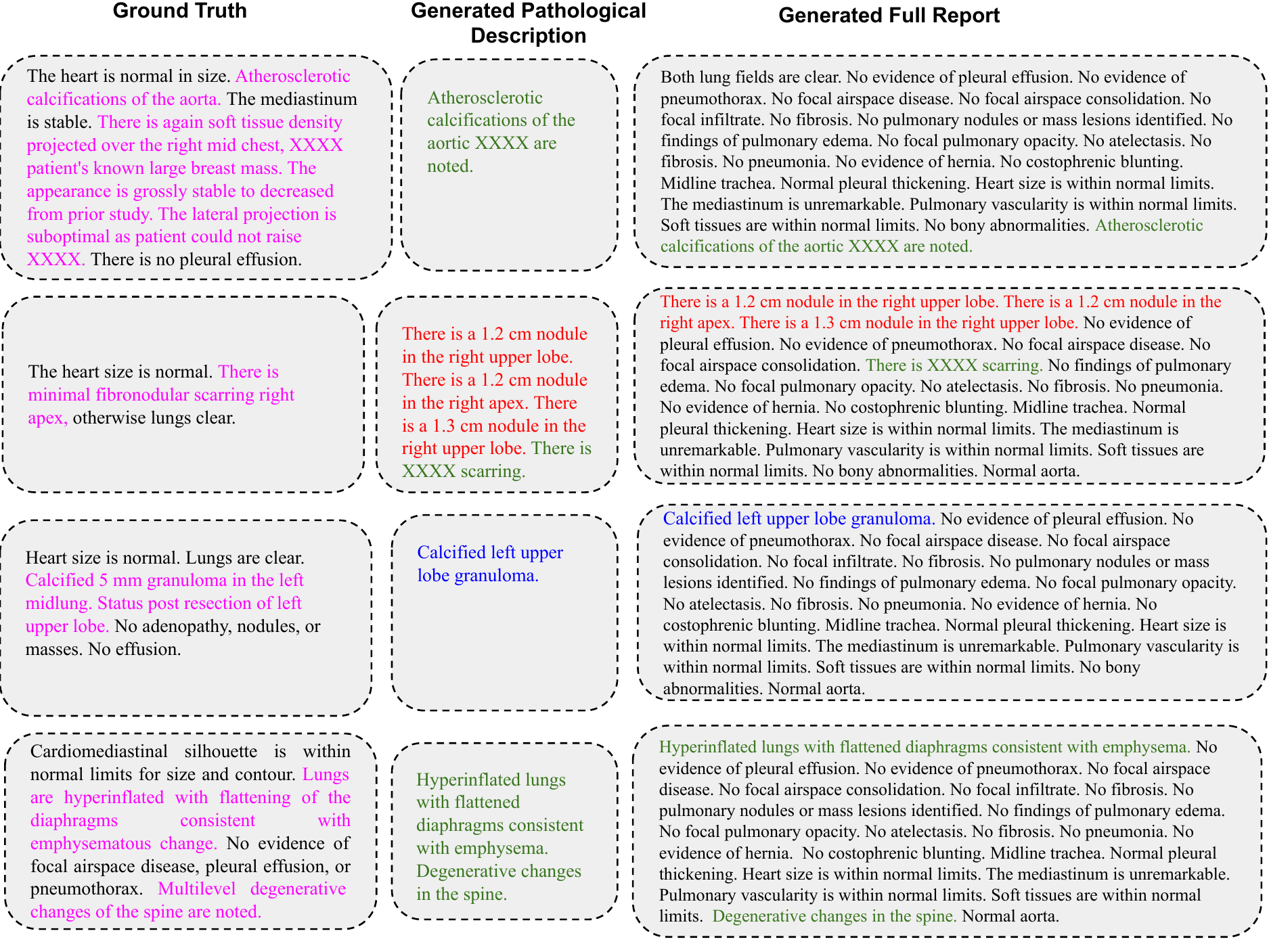}
\caption{Examples of the ground truth and the pathological descriptions and full reports generated using our approach The first column shows the findings from the IU X-RAY dataset. Abnormal findings in the original report are highlighted in {\color{magenta}{magenta}}. The second column shows the pathological description generated by our system. The third column shows the full report generated by our method. For both second and third column, correctly generated sentences are highlighted in {\color{green}{green}}, partially correct sentences are highlighted in {\color{blue}{blue}} and misleading sentences are highlighted in {\color{red}{red}}. Example 1 shows that the generated report is correct but missing some important information. Example 2 shows that the generated report is misleading. Example 3 shows that the generated report is correct but missing the measurements. Example 4 shows that the generated report is correct and reports all findings.}
\label{fig:error_analysis}
\end{figure*}

We provide 500 test samples from the test dataset and their corresponding generated reports to the domain expert for qualitative analysis. The model generated 78\% accurate reports, 12\% reports with missing information, and 10\% with misleading predictions. Further, these random samples were classified into \textit{normal} and \textit{abnormal} reports. Out of 183 normal reports, the model generated 99.95\% correct reports, 0.0\% reports with missing details, and 0.05\% misleading reports. Out of 317 abnormal reports, the model could produce 64\% accurate reports, 22\% of them with missing details, and 14.5\% with false reports. Table \ref{tab:human_eval} contains the results of the qualitative analysis.
Figure \ref{fig:error_analysis} shows the case studies of ground truth reports and generated pathological descriptions and full reports by our method for the above mentioned categories.

\section{Summary, Conclusion and Future Work}
We present a template-based approach for generating X-ray reports from radiographs. Our model generates small sentences exclusively for abnormalities, which are then substituted in the normal report template to produce a high-quality radiology report. We create a replacement dataset that contains pathological descriptions and their corresponding normal sentences from the normal report template. Our experimental results demonstrate that, compared to the State-of-the-Art models, the BLEU-1, ROUGE-L, METEOR, and CIDEr scores of the full reports generated by our approach are raised by 25\%, 36\%, 44\% and 48\%, respectively. Also, clinical evaluation metrics show that our method generates more clinically accurate reports than the State-of-the-Art methods. Unlike other State-of-the-Art models, our methodology does not put excessive emphasis on normal sentences. In the future, we plan to apply the proposed method to generate radiology reports for CT, MRI, \textit{etc}. For our experiments, we have used all samples from the IU X-RAY dataset. But we have used only 44578 reports out of 227827 reports for the MIMIC-CXR dataset. Our immediate plan is to perform experiments on the whole MIMIC-CXR dataset. Take away from our work is that creating smaller sentences of pathological descriptions and replacing them in the normal template produces better quality reports than generating the whole report at once.

\section*{Limitations}
Data unbalancing is one of the limitations of our work. In the future, we would like to address this problem by data oversampling or undersampling. For our experiments, we have used all samples from the IU X-ray dataset. But from the MIMIC-CXR dataset, we have used only 44578 reports out of 227827 reports. Our results for the MIMIC-CXR dataset might differ when we use the whole dataset. To evaluate the generated pathological descriptions, we consider the pathological descriptions that we extract from original reports as ground truth. To evaluate the generated full reports, we generate templated reports by replacing ground truth pathological description in normal report template and consider it as ground truth. So it considers the abnormalities from the original reports and the normal sentences from the normal report template. Automatic generation of chest X-ray reports will make it easier for radiologists to diagnose and write reports. Our model achieved comparable performance with State-of-the-Art models on chest X-ray report generation. In realistic scenarios, it is still a long way from being used clinically.

\section*{Ethics Statement}
The IU Chest X-ray dataset's authors used appropriate techniques to de-identify the text reports. Data is anonymized; hence, our model will not disclose information about the patient's identity. The MIMIC-CXR dataset does not contain any information related to the patient's identity, like name, age, or address. These reports are also anonymized. Data itself does not reveal the patient's identity; hence, our model also does not reveal the patient's identity.

\section*{Acknowledgements}
We thank the radiologist, Dr. Milind Gune, for his helpful suggestions and feedback. 

\bibliography{anthology,custom}
\bibliographystyle{acl_natbib}

\appendix

\end{document}